\documentclass[11pt]{article}
\usepackage{acl2016}
\usepackage{times}
\usepackage{latexsym}
\usepackage{amsmath}
\usepackage{amssymb}
\usepackage{gensymb}
\usepackage[pdftex]{graphicx}
\usepackage{color}
\DeclareGraphicsExtensions{.pdf,.jpeg,.jpg,.png}
\usepackage{adjustbox}
\usepackage{url}
\usepackage{multirow}
\usepackage{varwidth}

\aclfinalcopy 
\def\aclpaperid{286} 

\usepackage{algorithmic}
\usepackage{algorithm}

\title{Syntactically Guided Neural Machine Translation}

\author{Felix Stahlberg$^\dag$ \and Eva Hasler$^\dag$ \and Aurelien Waite$^\ddagger$ \and Bill Byrne$^{\ddagger\dag}$ \\
\\
    $^\dag$Department of Engineering, University of Cambridge, UK  \\
\\
    $^\ddagger$SDL Research, Cambridge, UK
}

\date{}

\begin{document}

\maketitle

\begin{abstract}
We investigate the use of hierarchical phrase-based SMT lattices in end-to-end neural machine translation (NMT).    Weight pushing transforms the Hiero scores for complete translation hypotheses, with the full translation grammar score and full n-gram language model score,  into posteriors compatible with NMT predictive probabilities.   With a slightly modified NMT beam-search decoder  we find gains over both Hiero and NMT decoding alone, with practical advantages in extending NMT to very large input and output vocabularies. 
\end{abstract}

\section{Introduction}
\label{sec:intro}

We report on investigations motivated by the idea that the structured search spaces defined by syntactic machine translation approaches such as Hiero~\cite{hierarchical} can be used to guide 
Neural Machine Translation (NMT)~\cite{kalchbrenner,sutskever,choPhraseRepresentation,bahdanau}.    
NMT and Hiero have complementary strengths and weaknesses and differ markedly in how they define probability distributions over translations and what search procedures they use.

The NMT encoder-decoder formalism  provides a  probability distribution over translations 
$\mathbf y=y_1^T$ of a source sentence $\mathbf x$ as~\cite{bahdanau}
\begin{equation}
P(y_1^T|\mathbf{x})  =\prod_{t=1}^{T}P(y_t|y_{1}^{t-1},\mathbf{x}) 
 = \prod_{t=1}^T g(y_{t-1}, s_t, c_t)
\label{eq:nmt}
\end{equation}
where $s_t = f(s_{t-1}, y_{t-1}, c_{t})$ is a decoder state variable and $c_t$ is a context vector depending on the source sentence and the attention mechanism.    

This posterior distribution is potentially very powerful, however it does not easily lend itself to sophisticated search procedures.    Decoding is done by `beam search to find a translation that approximately maximizes the conditional probability'~\cite{bahdanau}.   Search looks only one word ahead and no deeper than the  beam.

Hiero defines 
a synchronous context-free grammar (SCFG) with  rules: $X \rightarrow \left<\alpha, \gamma\right>$, where $\alpha$ and $\gamma$ are strings of terminals and non-terminals in the source and target languages.  A target language sentence $\mathbf y$ can be a translation of a source language sentence $\mathbf x$ if there is a derivation $D$ in the grammar which yields both ${\bf y}$ and ${\bf x}$: ${\bf y} = {\bf y}(D)$, ${\bf x}={\bf x}(D)$.  
This defines a regular language ${\mathcal Y}$ over strings in the target language via a projection of the  sentence to be translated: ${\mathcal Y} = \{ {\bf y}(D) : {\bf x}(D) = {\bf x} \}$~\cite{hifst-repr,J14-3008}.   
Scores are defined over derivations via a log-linear model with features $\{\phi_i\}$ and  weights $\lambda$. 
The  decoder searches for the translation ${\bf y}(D)$ in $\mathcal Y$ with the highest derivation score $S(D)$~\cite[Eq.~24]{hierarchical}~:

\begin{equation}
\hspace*{-1em}
\hat {\bf y} = {\bf y}\left( \operatorname*{argmax}_{D :  {\bf x}(D) = {\bf x} } \underbrace{P_G(D)  P_{LM} ({\bf y}(D))^{\lambda_{LM}} }_{S(D)}  \right)
\label{eq:hiero}
\end{equation}
where $P_{LM}$ is an n-gram language model and 
 $ P_G(D) \propto   \prod_{(X\rightarrow\left<\gamma, \alpha\right>) \in D}  \prod_i \phi_i (X\rightarrow \left<\gamma, \alpha\right>)^{\lambda_i} $.

Hiero decoders attempt to avoid search errors when combining the translation and language model for the translation hypotheses~\cite{hierarchical,hifst}.  These procedures search over a vast space of translations, much larger than is considered by the NMT beam search.   However the Hiero context-free grammars that make efficient search possible are weak models of translation.  The basic Hiero formalism can be extended through `soft syntactic constraints'~\cite{venugopal-EtAl:2009:NAACLHLT09,marton2008soft} or by adding very high dimensional features~\cite{chiang200911},  however the translation score assigned by the grammar is still only the product of probabilities of individual rules.   From the modelling perspective, this is an overly strong conditional independence assumption.  NMT clearly has the potential advantage in incorporating long-term context into translation scores.     

NMT and Hiero differ in how they `consume' source words.   Hiero applies the translation rules to the source sentence via the CYK algorithm, with each derivation yielding a complete and unambiguous translation of the source words.   The NMT beam decoder does not have an explicit mechanism for tracking source coverage, and there is evidence that may lead to both `over-translation' and `under-translation'~\cite{coverage-based}.    

NMT and Hiero also differ in their internal representations.    The NMT continuous representation captures morphological, syntactic and semantic similarity~\cite{collobert2008unified} across words and phrases.   However, extending these representations to the large vocabularies needed for open-domain MT is an open area of research~\cite{largeTargetVoc,unkReplace,bpe,chitnis-denero:2015:EMNLP}.    By contrast,  Hiero (and other symbolic systems) can easily use translation grammars and language models with very large vocabularies~\cite{Heafield-estimate,linMapReduceBook}.    Moreover, words and phrases can be easily added to a fully-trained symbolic MT system.    This is an important consideration for commercial MT, as customers often wish to customise and personalise SMT systems for their own application domain.   Adding new words and phrases to an NMT system is not as straightforward, and it is not clear that the advantages of the continuous representation can be extended to the new additions to the vocabularies.

NMT has the advantage of including long-range context in modelling individual translation hypotheses.  Hiero considers a much bigger search space, and can incorporate n-gram language models,  but a much weaker translation model.    
In this paper we try to exploit the strengths of each approach. We propose to guide NMT decoding using Hiero. We show that restricting the search space of the NMT decoder to a subset of ${\mathcal Y}$ spanned by Hiero effectively counteracts NMT modelling errors. This can be implemented by generating translation lattices with Hiero, which are then rescored by the NMT decoder. Our approach addresses the limited vocabulary issue in NMT as we replace NMT OOVs with lattice words from the much larger Hiero vocabulary.   We also find good gains from neural and Kneser-Ney n-gram language models.

\section{Syntactically Guided NMT (SGNMT)}
\label{sec:gnmt}

\subsection{Hiero Predictive Posteriors}
\label{ssec:hiero-predictive}

The Hiero decoder generates translation hypotheses as weighted finite state acceptors (WFSAs), or lattices, with weights in the tropical semiring.   For a translation hypothesis ${\bf y}(D)$ arising from the Hiero derivation $D$,   the path weight in the WFSA 
is $-\log S(D)$, after  Eq.~\ref{eq:hiero}.      While this representation is correct with respect to the Hiero translation grammar and language model scores,  having Hiero scores at the path level is not convenient for working with the NMT system.  What we need are  predictive probabilities in the form of Eq.~\ref{eq:nmt}.   

The Hiero WFSAs are determinised and minimised with epsilon removal under the tropical semiring, and  weights are pushed towards the initial state under the log semiring~\cite{fstpush}.    The resulting transducer is stochastic in the log semiring,  
i.e. the log sum of the arc log probabilities  leaving a state is $0~(=\log 1)$.    In addition,  because the WFSA is deterministic, there is a unique path leading to every state, which corresponds to a unique Hiero translation prefix.  Suppose a path to a state accepts the translation prefix $y_1^{t-1}$.   
An outgoing arc from that state with symbol $y$ has a weight that corresponds to the (negative log of the)  conditional probability 
\begin{equation}
P_{Hiero} (y_t =y | y_1^{t-1}, {\bf x}).
\label{eq:hierocp}
\end{equation}
This conditional probability is such that  for a Hiero translation ${y_1^T} = {\bf y}(D)$ accepted by the WFSA
\begin{equation}
P_{Hiero}(y_1^T) = \prod_{t=1}^T P_{Hiero}(y_t | y_1^{t-1}, {\bf x}) \propto S(D).
\end{equation}
The Hiero WFSAs have been transformed so that their arc weights have the negative log of the conditional probabilities defined in Eq.~\ref{eq:hierocp}.   All the probability mass of this distribution is concentrated on the Hiero translation hypotheses.    The complete translation and language model scores computed over the entire Hiero translations are pushed as far forward in the WFSAs as possible.  This is commonly done for left-to-right decoding in speech recognition~\cite{Mohri200269}.

\subsection{NMT--Hiero Decoding}
\label{ss:nmthiero}
As above,  suppose a path to a state in the WFSA accepts a Hiero translation 
prefix $y_1^{t-1}$, and let $y_t$ be a symbol on an outgoing arc from that state.   We define the joint NMT+Hiero score as 
\begin{multline}
\log  P(y_t | y_1^{t-1}, {\bf x})  = \\
  \lambda_{Hiero} \log P_{Hiero} (y_t | y_1^{t-1}, {\bf x}) \; + \\
   \lambda_{NMT} \left\{\begin{array}{@{\hspace{0.05em}}l@{\hspace{0.35em}}l@{\hspace{0.0em}}} 
   \log P_{NMT} (y_t| y_1^{t-1}, {\bf x}) & y_t \in \Sigma_{NMT} \\ 
   \log P_{NMT} ({\tt unk}| y_1^{t-1}, {\bf x}) & y_t \not\in \Sigma_{NMT}
   \end{array}\right.
   \label{eq:joint}
\end{multline}

Note that the \textsc{NMT-Hiero} decoder only considers hypotheses in the Hiero lattice. As discussed earlier,  the Hiero vocabulary can be much larger than the NMT output vocabulary $\Sigma_{NMT}$.    If a Hiero translation  contains a word not in the NMT vocabulary, the NMT model provides a score and updates its decoder state as for an unknown word.

\begin{table}[b!]
\small
\centering
\begin{tabular}{|@{\hspace{0.35em}}l@{\hspace{0.35em}}|@{\hspace{0.35em}}r@{\hspace{0.35em}}|@{\hspace{0.35em}}r@{\hspace{0.35em}}|@{\hspace{0.35em}}r@{\hspace{0.35em}}|@{\hspace{0.35em}}r@{\hspace{0.35em}}|@{\hspace{0.35em}}r@{\hspace{0.35em}}|@{\hspace{0.35em}}r@{\hspace{0.35em}}|}
\hline & \multicolumn{2}{@{\hspace{0.35em}}c@{\hspace{0.35em}}|@{\hspace{0.35em}}}{\bf Train set} & \multicolumn{2}{@{\hspace{0.35em}}c@{\hspace{0.35em}}|@{\hspace{0.35em}}}{\bf Dev set} & \multicolumn{2}{@{\hspace{0.35em}}c@{\hspace{0.35em}}|}{\bf Test set} \\
& en & de & en & de & en & de \\
 \hline
\# sentences & \multicolumn{2}{@{\hspace{0.35em}}c@{\hspace{0.35em}}|@{\hspace{0.35em}}}{4.2M} & \multicolumn{2}{@{\hspace{0.35em}}c@{\hspace{0.35em}}|@{\hspace{0.35em}}}{6k} & \multicolumn{2}{@{\hspace{0.35em}}c@{\hspace{0.35em}}|}{2.7k} \\
 \hline
\# word tokens & 106M & 102M & 138k & 138k & 62k & 59k \\
 \hline
\# unique words & 647k & 1.5M & 13k & 20k & 9k & 13k \\
 \hline
OOV (Hiero) & 0.0\% & 0.0\% & 0.8\% & 1.6\%  & 1.0\% & 2.0\% \\
 \hline
OOV (NMT) & 1.6\% & 5.5\% & 2.5\% & 7.5\% & 3.1\% & 8.8\% \\ \hline \hline
& en & fr & en & fr & en & fr \\
 \hline
\# sentences & \multicolumn{2}{@{\hspace{0.35em}}c@{\hspace{0.35em}}|@{\hspace{0.35em}}}{12.1M} & \multicolumn{2}{@{\hspace{0.35em}}c@{\hspace{0.35em}}|@{\hspace{0.35em}}}{6k} & \multicolumn{2}{@{\hspace{0.35em}}c@{\hspace{0.35em}}|}{3k} \\
 \hline
\# word tokens & 305M & 348M & 138k & 155k & 71k & 81k \\
 \hline
\# unique words & 1.6M & 1.7M & 14k & 17k & 10k & 11k \\
 \hline
OOV (Hiero) & 0.0\% & 0.0\% & 0.6\% & 0.6\%  & 0.4\% & 0.4\% \\
 \hline
OOV (NMT) & 3.5\% & 3.8\% & 4.5\% & 5.3\% & 5.0\% & 5.3\% \\
\hline
\end{tabular}
\caption{\label{tab:wmt-data} Parallel texts and vocabulary coverage on {\em news-test2014}. }
\end{table}

Our decoding algorithm 
is a natural extension of beam search decoding for NMT. Due to the form of Eq.~\ref{eq:joint} we can build up hypotheses from left-to-right on the target side. Thus, we can represent a partial hypothesis $h=(y_1^t,h_s)$ by a translation prefix $y_1^t$ and an accumulated score $h_s$. At each iteration we extend the current hypotheses by one target token, 
until the best scoring hypothesis reaches a final state of the Hiero lattice.
We refer to this step as {\em node expansion}, and in Sec.~\ref{ssec:sgnmt-performance} we report the number of node expansions per sentence, as an indication of computational cost.

We can think of the decoding algorithm as breath-first search through the translation lattices with a limited number of active hypotheses (a beam).  Rescoring is done on-the-fly: as the decoder traverses an edge in the WFSA, we update its weight by Eq.~\ref{eq:joint}.   The output-synchronous characteristic of beam search enables us to compute the NMT posteriors only once for each history based on previous calculations.

Alternatively, we can think of the algorithm as NMT decoding with revised posterior probabilities:  instead of selecting the most likely symbol $y_t$ according the NMT model, we adjust the NMT posterior with the Hiero posterior scores and delete NMT entries that are not allowed by the lattice. This may result in NMT choosing a different symbol, which is then fed back to the neural network for the next decoding step.

\begin{table*}[t!]
\small
\centering
\begin{minipage}[t]{1\columnwidth}
\centering
\begin{tabular}{|l|r|l|r|}
\hline
\multicolumn{2}{|c|}{\textbf{\cite[Tab.~2]{largeTargetVoc}}} & \multicolumn{2}{|c|}{\textbf{SGNMT}}\\
\hline
\textbf{Setup} & \textbf{BLEU} & \textbf{Setup} & \textbf{BLEU} \\
\hline
\textsc{Basic NMT} & 16.46 & \textsc{Basic NMT} & 16.31 \\
NMT-LV & 16.95 & \textsc{Hiero} & 19.44 \\
+ UNK Replace & 18.89 & \textsc{NMT-Hiero} & 20.69\\
-- & -- & + Tuning & 21.43 \\
+ Reshuffle & 19.40 & + Reshuffle & 21.87\\
\hline
+ Ensemble & 21.59 & \multicolumn{2}{|c}{} \\
\cline{1-2}
\end{tabular}

\hspace{1.5cm}

(a) English-German

\end{minipage}
\begin{minipage}[t]{1\columnwidth}
\centering
\begin{tabular}{|l|r|l|r|}
\hline
\multicolumn{2}{|c|}{\textbf{\cite[Tab.~2]{largeTargetVoc}}} & \multicolumn{2}{|c|}{\textbf{SGNMT}}\\
\hline
\textbf{Setup} & \textbf{BLEU} & \textbf{Setup} & \textbf{BLEU} \\
\hline
\textsc{Basic NMT} & 29.97 & \textsc{Basic NMT} & 30.42 \\
NMT-LV & 33.36 & \textsc{Hiero} & 32.86 \\
+ UNK Replace & 34.11 & \textsc{NMT-Hiero} & 35.37\\
-- & -- & + Tuning & 36.29 \\
+ Reshuffle & 34.60 & + Reshuffle & 36.61\\
\hline
+ Ensemble & 37.19 & \multicolumn{2}{|c}{} \\
\cline{1-2}
\end{tabular}

\hspace{1.5cm}

(b) English-French

\end{minipage}
\caption{\label{tab:wmt-results} BLEU scores on {\em news-test2014} calculated with \texttt{multi-bleu.perl}. \textsc{NMT-LV} refers to the \textsc{RNNsearch-LV} model from \cite{largeTargetVoc} for large output vocabularies. 
}
\end{table*}

\begin{table*}[t!]
\small
\centering

\begin{tabular}{@{\hspace{0em}}r@{\hspace{0.2em}}|c|c|c|c|c|c||r||r|r|@{\hspace{0em}}}
\cline{2-10}
& \textbf{Search} & \textbf{Vocab.} & \textbf{NMT} & \textbf{Grammar} & \textbf{KN-LM} & \textbf{NPLM} & \multicolumn{1}{|c||}{\textbf{\# of node exp-}} & \multicolumn{1}{|c|}{\textbf{BLEU}} & \multicolumn{1}{|c|}{\textbf{BLEU}} \\ 
 & \textbf{space} & & \textbf{scores} & \textbf{scores} & \textbf{scores} & \textbf{scores} & \multicolumn{1}{|c||}{\textbf{ansions per sen.}} & \textbf{(single)} & \textbf{(ensemble)} \\\cline{2-10}
{\scriptsize 1} & Lattice & Hiero &  & $\checkmark$ & $\checkmark$ & & -- & \multicolumn{2}{|c|}{21.1 (Hiero)} \\  
{\scriptsize 2} & Lattice & Hiero & & $\checkmark$ & $\checkmark$ & $\checkmark$ & -- & \multicolumn{2}{|c|}{21.7 (Hiero)} \\  
\cline{2-10}
{\scriptsize 3} & Unrestricted & NMT & $\checkmark$ & & & & 254.8 & 19.5 & 21.8  \\  
\cline{2-10}
{\scriptsize 4} & 100-best & Hiero & $\checkmark$ & & & & \multirow{3}{*}{\begin{varwidth}{7em}
\begin{flushright}
2,233.6\\
(DFS: 832.1)
\end{flushright}
    \end{varwidth}} & 22.8 & 23.3  \\  
{\scriptsize 5} & 100-best & Hiero & $\checkmark$ & $\checkmark$ & $\checkmark$ & &  & 22.9 & 23.4 \\  
{\scriptsize 6} & 100-best & Hiero & $\checkmark$ & $\checkmark$ & $\checkmark$ & $\checkmark$ &  & 22.9 & 23.3  \\  
\cline{2-10}
{\scriptsize 7} & 1000-best & Hiero & $\checkmark$ & & & & \multirow{3}{*}{\begin{varwidth}{7em}
\begin{flushright}
21,686.2\\
(DFS: 6,221.8)
\end{flushright}
    \end{varwidth}} & 23.3 & 23.8  \\  
{\scriptsize 8} & 1000-best & Hiero & $\checkmark$ & $\checkmark$ & $\checkmark$ & & & 23.4 & 23.9  \\  
{\scriptsize 9} & 1000-best & Hiero & $\checkmark$ & $\checkmark$ & $\checkmark$ & $\checkmark$ & & 23.5 & 24.0  \\  
\cline{2-10}
{\scriptsize 10} & Lattice & NMT & $\checkmark$ & & & & 243.3 & 20.3 & 21.4 \\ 
{\scriptsize 11} & Lattice & Hiero & $\checkmark$ & & & & 243.3 & 23.0 & 24.2 \\  
{\scriptsize 12} & Lattice & Hiero & $\checkmark$ & $\checkmark$ & & & 243.3 & 23.0 & 24.2  \\  
{\scriptsize 13} & Lattice & Hiero & $\checkmark$ & & $\checkmark$ & & 240.5 & 23.4 & 24.5 \\  
{\scriptsize 14} & Lattice & Hiero & $\checkmark$ & $\checkmark$ & $\checkmark$ & & 243.9 & 23.4 & 24.4 \\  
{\scriptsize 15} & Lattice & Hiero & $\checkmark$ & $\checkmark$ & $\checkmark$ & $\checkmark$ & 244.3 & 24.0 & 24.4  \\  
\cline{2-10}
{\scriptsize 16} & \multicolumn{7}{|l||}{Neural MT -- UMontreal-MILA~\cite{montreal15}}  & 22.8 & 25.2 \\
\cline{2-10}
\end{tabular}
\caption{\label{tab:wmt15-results} BLEU English-German {\em news-test2015}   scores calculated with \texttt{mteval-v13a.pl}. }
\end{table*}

\section{Experimental Evaluation}
We evaluate SGNMT on the WMT {\em news-test2014} test sets (the filtered version) for English-German (En-De) and English-French (En-Fr).  
We also report results on WMT {\em news-test2015} En-De.

The En-De training set includes  {\em Europarl v7},  {\em Common Crawl}, and  {\em News Commentary v10}. 
Sentence pairs with sentences longer than 80 words or length ratios exceeding  2.4:1 were deleted, as were   {\em Common Crawl} sentences from other languages~\cite{langDetect}. 
The En-Fr NMT system was trained on preprocessed data~\cite{fr-data} used by previous work~\cite{sutskever,bahdanau,largeTargetVoc}, but with truecasing like our Hiero baseline.  
Following \cite{largeTargetVoc}, we use  {\em news-test2012} and {\em news-test2013} as a development set. 
The NMT vocabulary size is 50k for En-De and 30k for En-Fr, taken as the most frequent words in training~\cite{largeTargetVoc}.
Tab.~\ref{tab:wmt-data} provides statistics  and shows the severity of the OOV problem for NMT.

The \textsc{Basic NMT} system is built using the  Blocks framework~\cite{blocks} based on the Theano library~\cite{theano} with standard hyper-parameters~\cite{bahdanau}: the encoder and decoder networks consist of 1000~gated recurrent units~\cite{choPhraseRepresentation}. The decoder uses a single maxout~\cite{maxout} output layer with the feed-forward attention model~\cite{bahdanau}.

The En-De Hiero system uses rules which encourage verb movement~\cite{hifst-grammar}. The rules for En-Fr were extracted from the full data set available at the WMT'15 website using a shallow-1 grammar~\cite{hifst-grammar}. 5-gram Kneser-Ney language models (KN-LM) for the Hiero systems were trained on  WMT'15 parallel and monolingual data~\cite{Heafield-estimate}.  Our SGNMT system\footnote{http://ucam-smt.github.io/sgnmt/html/} is built with the Pyfst interface~\footnote{https://pyfst.github.io/} to OpenFst~\cite{openfst}.

\subsection{SGNMT  Performance}
\label{ssec:sgnmt-performance}

Tab.~\ref{tab:wmt-results} compares our combined NMT+Hiero decoding with NMT results in the literature. We use a beam size of 12.  In En-De and in En-Fr,  we find that our \textsc{Basic NMT} system performs similarly (within 0.5 BLEU) to previously published results (16.31 vs. 16.46 and 30.42 vs. 29.97).

In  \textsc{NMT-Hiero}, decoding is as described in Sec.~\ref{ss:nmthiero},  but with $ \lambda_{Hiero} = 0$.  The decoder searches through the Hiero lattice, ignoring the Hiero scores, but using Hiero word hypotheses in place of any UNKs that might have been produced by \textsc{NMT}.   The results show that \textsc{NMT-Hiero} is much more effective in fixing NMT OOVs than the `UNK Replace' technique~\cite{unkReplace}; this holds in both En-De and En-Fr.    

For the  \textsc{NMT-Hiero+Tuning} systems, lattice MERT~\cite{lmert} is used to optimise $\lambda_{Hiero}$ and $\lambda_{NMT}$ on the tuning sets.   
This yields further gains in both En-Fr and En-De,  suggesting that in addition to fixing UNKs, the Hiero predictive posteriors  can be used to improve the NMT translation model scores.

Tab.~\ref{tab:wmt15-results} reports results of our En-De system with reshuffling and tuning on {\em news-test2015}.  BLEU scores are directly comparable to  WMT'15 results~\footnote{http://matrix.statmt.org/matrix/systems\_list/1774}. By comparing row 3 to row 10, we see that constraining NMT to the search space defined by the Hiero lattices yields an improvement of +0.8 BLEU for single NMT. If we allow Hiero to fix NMT UNKs, we see a further +2.7 BLEU gain (row 11). The majority of gains come from fixing UNKs, but there is still improvement from the constrained search space for single NMT.

\begin{figure}[!t]
\centering
\includegraphics[width=1\linewidth]{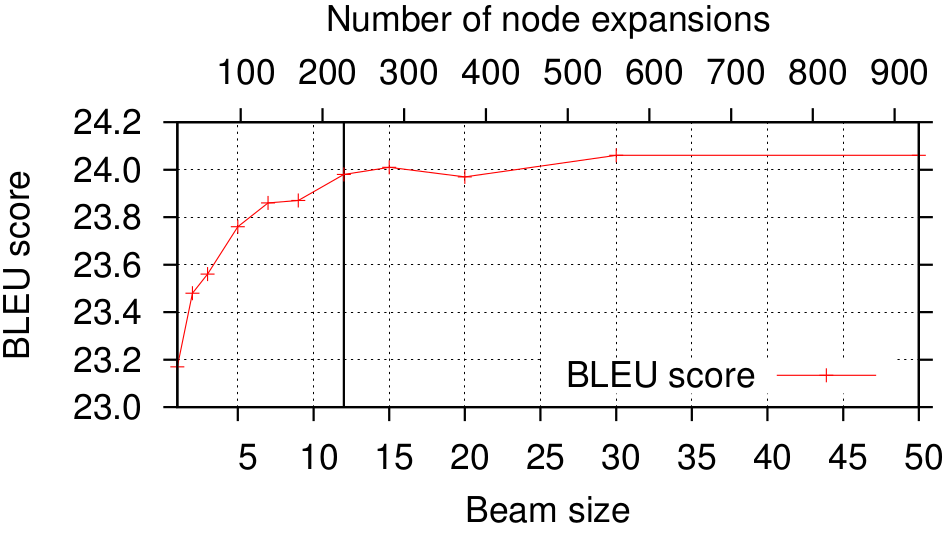}
\caption{Performance with NPLM over beam size on English-German {\em news-test2015}. A beam of 12 corresponds to row 15 in Tab.~\ref{tab:wmt15-results}.}
\label{fig:beam}
\end{figure}

We next investigate the contribution of the Hiero system scores. We see that, once lattices are generated, the KN-LM contributes more to rescoring than the Hiero grammar scores (rows 12-14). Further gains can be achieved by adding a feed-forward neural language model with NPLM \cite{nplm} (row 15). We observe that $n$-best list rescoring with NMT~\cite{neubigAtwat2015} also outperforms both the Hiero and NMT baselines, although lattice rescoring gives the best results (row 9 vs.\ row 15). Lattice rescoring with SGNMT also uses far fewer node expansions per sentence. We report $n$-best rescoring speeds for rescoring each hypothesis separately, and a depth-first (DFS) scheme that efficiently traverses the $n$-best lists. Both these techniques are very slow compared to lattice rescoring. Fig.~\ref{fig:beam} shows that we can reduce the beam size from 12 to 5 with only a minor drop in BLEU. This is nearly 100 times faster than DFS over the 1000-best list.

\begin{table}
\small
\centering
\begin{tabular}{|c|c|c|r|}
\hline
{\bf Determini-} & {\bf Minimi-} & {\bf Weight} & {\bf Sentences} \\
{\bf sation} & {\bf sation} & {\bf pushing} & {\bf per second} \\
\hline
$\checkmark$ & & & 2.51 \\
$\checkmark$ & $\checkmark$ & & 1.57 \\
$\checkmark$ & $\checkmark$ & $\checkmark$ & 1.47 \\
\hline
\end{tabular}
\caption{\label{tab:lattice-preprocessing} Time for lattice preprocessing operations on English-German {\em news-test2015}. }
\end{table}

\paragraph{Cost of Lattice Preprocessing}  As described in Sec.~\ref{ssec:hiero-predictive}, we applied determinisation, minimisation, and weight pushing to the Hiero lattices in order to work with probabilities. Tab.~\ref{tab:lattice-preprocessing} shows that those operations are generally fast\footnote{Testing environment: Ubuntu 14.04, Linux 3.13.0, single Intel\textsuperscript{\textregistered} Xeon\textsuperscript{\textregistered} X5650 CPU at 2.67 GHz}.

\begin{figure}[!t]
\centering
\includegraphics[width=0.76\linewidth]{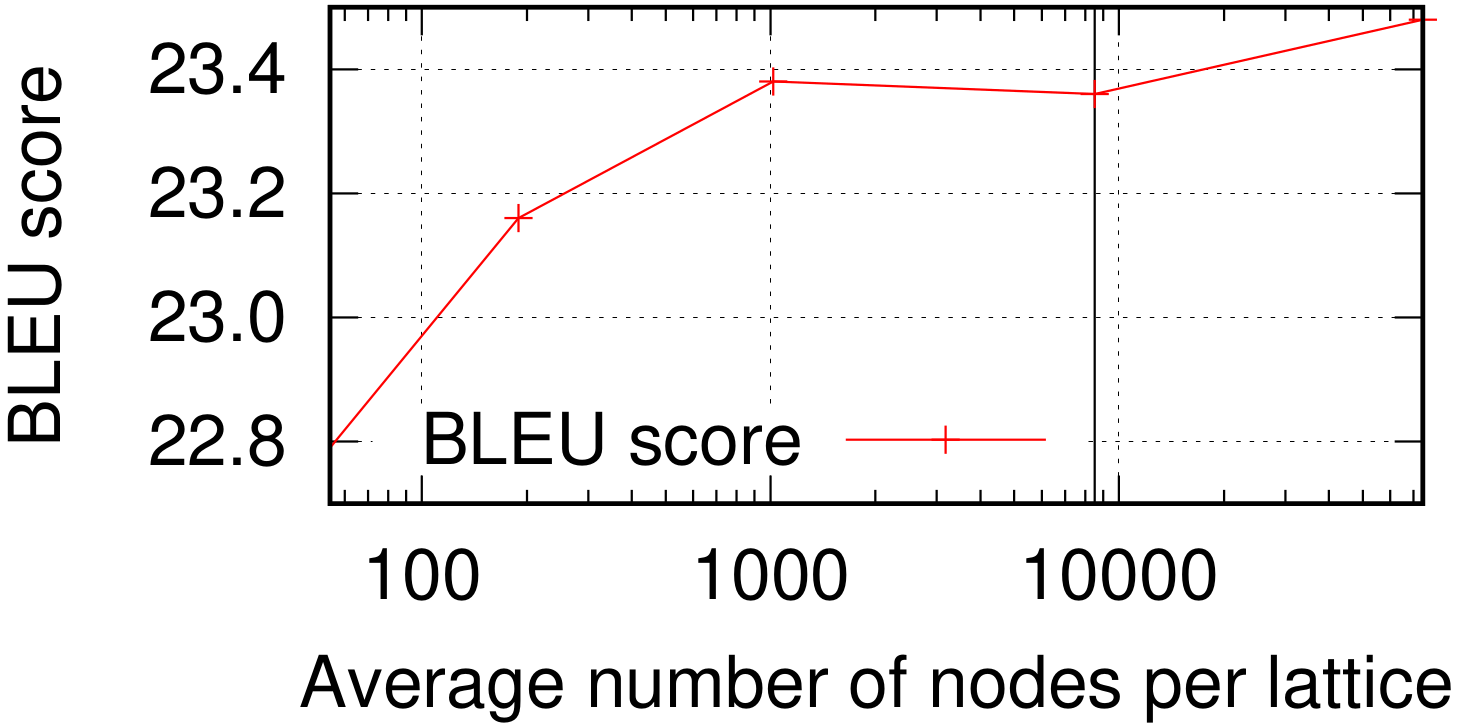}
\caption{SGNMT performance over lattice size on English-German {\em news-test2015}. 8,510 nodes per lattice corresponds to row 14 in Tab.~\ref{tab:wmt15-results}.}
\label{fig:pruning}
\end{figure}

\paragraph{Lattice Size} For previous experiments we set the Hiero pruning parameters such that lattices had 8,510 nodes on average. Fig.~\ref{fig:pruning} plots the BLEU score over the lattice size. We find that SGNMT works well on lattices of moderate or large size, but pruning lattices too heavily has a negative effect as they are then too similar to Hiero first best hypotheses. We note that lattice rescoring involves nearly as many node expansions as unconstrained NMT decoding. This confirms that the lattices at 8,510 nodes are already large enough for SGNMT.

\paragraph{Local Softmax}   
In SGNMT decoding we have the option of normalising the NMT translation probabilities over the words on outgoing words from each state rather than over the full 50,000 words translation vocabulary.   There are $\sim$4.5  arcs per state in our En-De'14 lattices, and so avoiding the full softmax could cause significant computational savings.   We find this leads to only a modest 0.5 BLEU degradation:  
 21.45 BLEU in En-De'14,  compared to 21.87 BLEU using NMT probabilities computed over the full vocabulary.

\paragraph{Modelling Errors vs.\ Search Errors}  In our En-De'14 experiments with $\lambda_{Hiero}=0$ we find that constraining the NMT decoder to the Hiero lattices yields translation hypotheses with much lower NMT probabilities than unconstrained \textsc{Basic NMT} decoding:  under the NMT model,  \textsc{NMT} hypotheses are 8,300 times more likely (median) than \textsc{NMT-Hiero} hypotheses.    We conclude (tentatively) that \textsc{Basic NMT} is not suffering only from search errors,  but rather that \textsc{NMT-Hiero}  discards some hypotheses ranked highly by the NMT model but lower in the evaluation metric.

\section{Conclusion}
We have demonstrated a viable approach to Syntactically Guided Neural Machine Translation formulated to exploit the rich, structured search space generated by Hiero and the long-context translation scores of NMT.    SGNMT does  not suffer from the severe limitation in vocabulary size of basic NMT and avoids any difficulty of extending distributed word representations to new vocabulary items not seen in training data.

\section*{Acknowledgements}

This work was supported in part by the U.K. Engineering and Physical Sciences Research Council (EPSRC grant EP/L027623/1).

\bibliography{refs}
\bibliographystyle{acl2016}

\end{document}